\documentclass[letterpaper]{article}
\usepackage{iccc}

\usepackage[usenames,dvipsnames,table]{xcolor}
\usepackage{colortbl}
\usepackage{times}
\usepackage{helvet}
\usepackage{courier}
\usepackage{graphicx}
\usepackage{balance}

\pdfoutput=1

\usepackage{times}

\usepackage{epsfig}
\usepackage{graphicx}
\usepackage[belowskip=-10pt,aboveskip=2pt,font=small]{caption}
\usepackage{dblfloatfix}

\usepackage{amsmath}
\usepackage{nccmath}
\usepackage{amssymb}

\usepackage{multirow}
\usepackage{rotating}
\usepackage{booktabs}
\usepackage{algorithm}

\usepackage{enumerate}
\usepackage[olditem,oldenum]{paralist}

\usepackage{mysymbols}

\usepackage[pagebackref=true,breaklinks=true,letterpaper=true,colorlinks,bookmarks=false,citecolor=blue,linkcolor=green]{hyperref}
\usepackage{xcolor}

\definecolor{orange}{rgb}{1,0.5,0}
\definecolor{lightsalmonpink}{rgb}{1.0, 0.6, 0.6}
\definecolor{verylightsalmonpink}{rgb}{0.966, 0.805, 0.797}
\definecolor{lightblue}{rgb}{0.862, 0.906, 0.984}
\definecolor{lightyellow}{rgb}{1.0, 0.945, 0.797}
\definecolor{lightgreen}{rgb}{0.835, 0.91, 0.828}
\definecolor{lightpurple}{rgb}{0.879, 0.832, 0.902}

\usepackage[utf8]{inputenc} 
\usepackage[T1]{fontenc}    
\usepackage{amsfonts}       
\usepackage{nicefrac}       
\usepackage{microtype}      
\usepackage{subcaption}
\usepackage[font=footnotesize,labelfont=footnotesize]{caption}
\usepackage{comment}

\newcommand{\rulesep}{\unskip\ \vrule width 2pt\ }
\newcommand{\themeall}[0]{\colorbox{lightblue}{\textsc{theme-all}}}
\newcommand{\themesome}[0]{\colorbox{lightgreen}{\textsc{theme-some}}}
\newcommand{\nothemesome}[0]{\colorbox{lightpurple}{\textsc{notheme-some}}}
\newcommand{\nothemeall}[0]{\colorbox{verylightsalmonpink}{\textsc{notheme-all}}}
\newcommand{\fontall}[0]{\colorbox{lightyellow}{\textsc{font}}}

\newcommand\numberthis{\addtocounter{equation}{1}\tag{\theequation}}

\newcommand{\skeleton}[0]{\includegraphics[width=.027\textwidth]{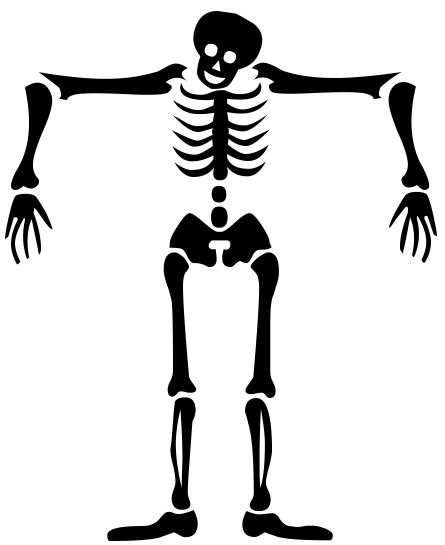}}
\newcommand{\rletter}[0]{\includegraphics[width=.0255\textwidth]{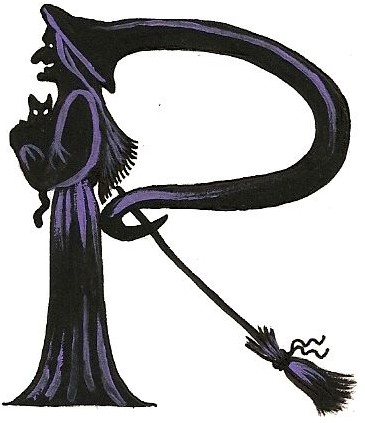}}
\newcommand{\aletter}[0]{\includegraphics[width=.03\textwidth]{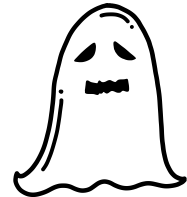}}

\newcommand{\namedref}[2]{\hyperref[#2]{#1~\ref*{#2}}}

\newcommand{\treat}[0]{TReAT}

\newcommand{\figureref}[1]{\namedref{Figure}{fig:#1}}

\newcommand{\code}[1]{\texttt{#1}}

\title{Trick or \skeleton \rletter e\aletter\skeleton : Thematic Reinforcement for Artistic Typography}
\author{
    Purva Tendulkar$^1$ \hspace{1.5pc}
    Kalpesh Krishna$^2$ \hspace{1.5pc}
	Ramprasaath R. Selvaraju$^1$ \hspace{1.5pc}
    Devi Parikh$^1$ \\
    $^1$Georgia Institute of Technology,~~
	$^2$University of Massachusetts, Amherst\\
	$^1${\tt\small \{purva, ramprs, parikh\}@gatech.edu},~~
    $^2${\tt\small kalpesh@cs.umass.edu}
}

\begin{document}
\maketitle
\begin{abstract}
\begin{quote}

An approach to make text visually appealing and memorable is \emph{semantic reinforcement} -- the use of visual cues alluding to the context or theme in which the word is being used to reinforce the message (e.g., Google Doodles). 
We present a computational approach for semantic reinforcement called \treat~ -- Thematic Reinforcement for Artistic Typography. 
Given an input word (e.g. \code{exam}) and a theme (e.g. \code{education}), the individual letters of the input word are replaced by cliparts relevant to the theme which visually resemble the letters -- adding creative context to the potentially boring input word. 
We use an unsupervised approach to learn a latent space to represent letters and cliparts and compute similarities between the two. 
Human studies show that participants can reliably recognize the word as well as the theme in our outputs (\treat s) and find them more creative compared to meaningful baselines.
\end{quote}
\end{abstract}

\section{Introduction}
\label{sec:introduction}

\begin{figure}[t]
\centering
\includegraphics[scale=0.1855]{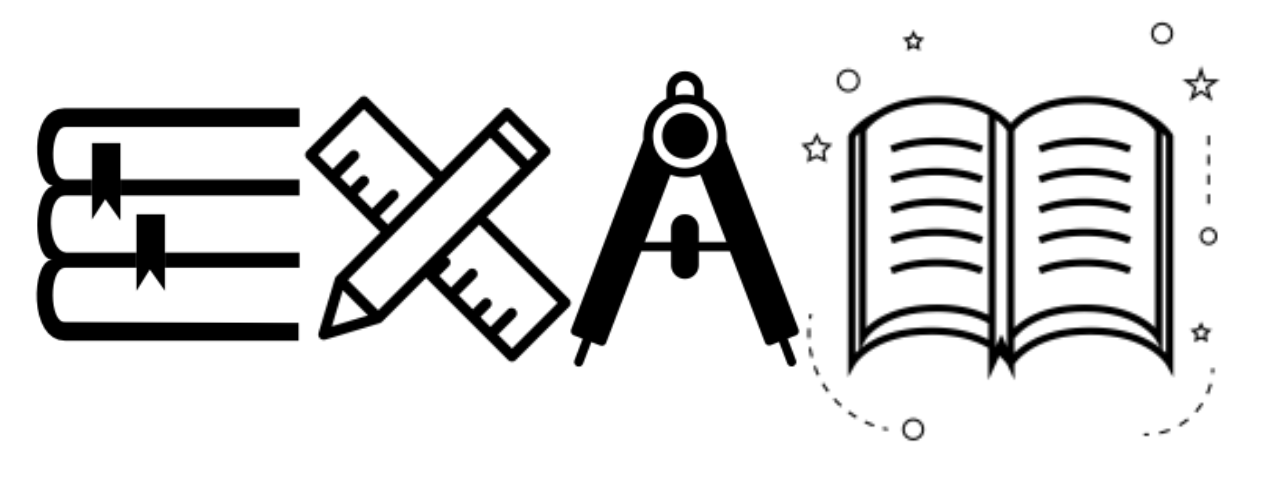}
\caption{A sample doodle (that we call \treat) generated by our system for the input word \code{exam} and theme \code{education}\protect \footnotemark.}
\label{fig:exam}
\end{figure}

We address the task of theme-based word typography: given a word (e.g., \code{exam}) and a theme (e.g., \code{education}), the task is to \textit{automagically} produce a doodle for the word in that theme as seen in \figureref{exam}. 
Concretely, the task is to replace each letter in the input word with a clipart from the input theme to produce a doodle, such that the word and theme can be easily identified from the doodle. 
Solving this task would be of value to a variety of creative applications such as stylizing text in advertising, designing logos -- essentially any application where a message needs to be conveyed to an audience in an effective and concise manner.

\begin{figure*}[ht]
	\centering
	\begin{subfigure}[t]{0.16\textwidth}
	    \centering
		\raisebox{0.2cm}{\includegraphics[width=0.9\textwidth]{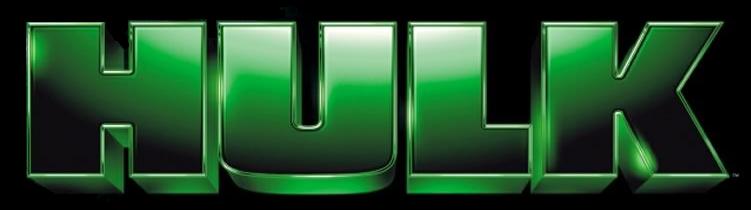}}
		\vspace{0.2cm}
        \caption{}
        \label{fig:font_color}
	\end{subfigure}
	\begin{subfigure}[t]{0.16\textwidth}
		\centering
		\includegraphics[width=0.9\textwidth]{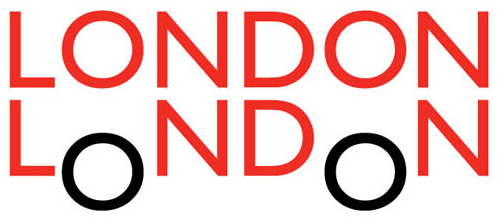}
		\vspace{0.2cm}
        \caption{}
        \label{fig:position}
	\end{subfigure}
	\begin{subfigure}[t]{0.16\textwidth}
		\centering
		\includegraphics[width=0.7\textwidth]{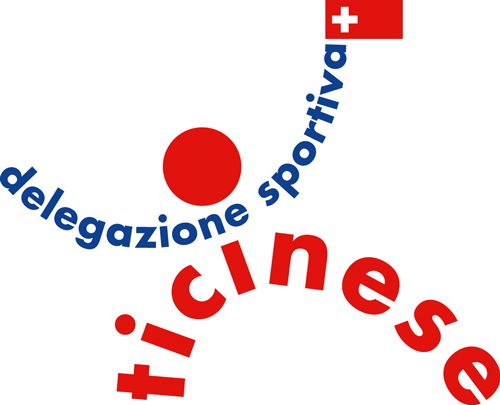}
		\vspace{0.2cm}
        \caption{}
        \label{fig:arrangement}
	\end{subfigure}
	\begin{subfigure}[t]{0.16\textwidth}
		\centering
		\includegraphics[width=0.7\textwidth]{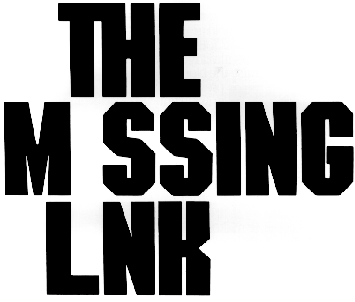}
		\vspace{0.2cm}
        \caption{}
        \label{fig:exclusion}
	\end{subfigure}
	\begin{subfigure}[t]{0.16\textwidth}
		\centering
        \includegraphics[width=0.8\textwidth]{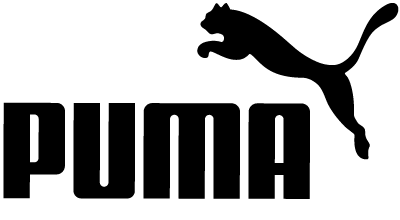}
       \vspace{0.2cm}
        \caption{}
        \label{fig:addition}
	\end{subfigure}
	\begin{subfigure}[t]{0.16\textwidth}
		\centering
        \includegraphics[width=0.9\textwidth]{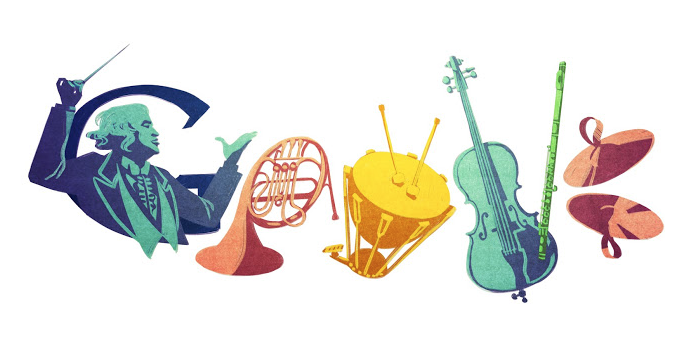}
        \vspace{0.2cm}
		\caption{}
        \label{fig:replacement}
	\end{subfigure}
    \vspace{10pt}
	\caption[]{Different methods for semantic reinforcement. a) font and color variations b) positioning of letters relative to each other c) arrangement of letters in a specific shape or direction d) exclusion of some letters e) addition of icons near letters f) replacement of letters. In this work we focus on f), semantic reinforcement via replacement. \protect \footnotemark}
    \label{fig:semantic_reinforcement}
\end{figure*}

Using graphic elements to emphasize the meaning of a word in reference to a related theme is referred to in graphic design as \textit{semantic reinforcement}.
This can be achieved in a number of ways, e.g., using different fonts and colors (\figureref{font_color}), changing the position of letters relative to one another (\figureref{position}), arranging letters in a specific direction or shape (\figureref{arrangement}), excluding some letters (\figureref{exclusion}), adding icons near or around the letters (\figureref{addition}), or replacing letters with icons (\figureref{replacement}). 
In our work, we focus on this last type, i.e., semantic reinforcement via replacement.

\footnotetext{Unless stated otherwise, all cliparts in the paper have been taken from The Noun Project - \url{https://thenounproject.com/}. The Noun Project contains cliparts created by different graphic designers on a variety of themes.}

This is a challenging task even for humans. 
It not only requires domain-specific knowledge for identifying a set of relevant cliparts to choose from, but also requires creative abilities to be able to visualize a letter in a clipart, and choose the best clipart for representing it.

The latter alone is challenging to automate -- both from a training and evaluation perspective. 
Training a model to automatically match letters to graphics is challenging because there is a lack of large-scale text-graphic paired datasets in each domain that might be of interest (e.g., clipart, logogram). 
Evaluation and thus iterative development of such models is also challenging because of subjectivity and inter-human disagreement on which letter resembles which graphic.

We present a computational approach -- Thematic Reinforcement of Artistic Typography (\treat) -- to generate doodles (\treat s) for semantic reinforcement of text via replacement. 
We represent letters in different fonts and cliparts from the Noun Project. 
in a common latent space. 
These latent representations are learned such that they have two characteristics: 
(1) The letters can be correctly recognized (e.g., \textit{a} vs. \textit{b}) in the latent space and 
(2) The letters and cliparts can be reconstructed accurately from the latent space. 
A reconstruction loss ensures that letters and clipart that are close in the latent space also look similar in the image space. 
A classification loss ensures that the latent space is informed by discriminative features that make one letter different from the other. 
This allows us to match cliparts to letters in a way that preserves distinctive visual features of the letters, making it easier for humans to identify the letter being depicted by a clipart. 
At test time, given a word and a theme as input, we first retrieve cliparts from the Noun Project that match that theme. For each letter in the word, we find the theme-relevant clipart which minimizes the distance from it across a variety of fonts. 
If the distance is low enough, we replace the letter with the clipart.

We run human studies to show that subjects can reliably recognize the word as well as the theme from our \treat s, and find them creative relative to meaningful baselines.

\footnotetext{Examples a) to e) were taken from \href{https://graphicdesign.stackexchange.com/a/115356/132602}{this answer} on StackExchange. Example f) is a Google Doodle.}

Our contributions are as follows:
\begin{itemize}
 \item We present \treat~-- an unsupervised approach to creatively stylize a word using theme-based cliparts.\footnote{Our code will be released on GitHub.}
  \item We define a set of evaluation metrics for this task and evaluate our approach under these metrics.
  \item We show that our approach outperforms meaningful baselines in terms of word recognition, theme recognition, as well as creativity.
\end{itemize}

\section{Related work}
\label{sec:related_work}

Early human communication was through symbols and hieroglyphs \cite{frutiger1989signs}, \cite{schmandt2014evolution}. 
This involved the use of characters to represent an entire word, phrase or concept. 
Then language evolved and we started using the alphabet for creation of new words to represent concepts. 
However many languages (e.g. Chinese and Japanese) still make use of pictograms or logograms to depict specific words. 
Today, symbols and logos are used for creative applications to increase the communication bandwidth -- to convey abstract concepts, express rich emotions, or reinforce messages \cite{shiojiri2013visual}, \cite{clawson2012using}, \cite{takasaki2007design}. 
Our work produces a visual depiction of text by reasoning about similarity between the visual appearance of a letter and clipart imagery. 
We describe prior work in each of these domains: creativity through imagery and creativity through visual appearance of text.

\subsection{Creativity through imagery}
There is previous work on evoking emotional responses through the modification of images. 
Work on visual blending of emojis combines different concepts to create novel emojis~\cite{martins2018how}. 
Visual blending has also been explored for combining two animals to create images depicting fictional hybrid animals~\cite{martins2015good}. 
Our approach tries to induce creativity by entirely replacing a letter with a clipart. 
Towards the end of the paper we briefly describe a generative modeling approach we experiment with. 
That can be thought of as blending between a letter and a clipart. 
However, the motivation behind our blends is different in that we intend to generate an output that looks like both the clipart and letter, as opposed to blending specific local elements of both images to create a new concept.

Work on Vismantic \cite{xiaovismantic} represents abstract concepts visually by combining images using juxtaposition, fusion, and replacement. 
Our work also represents a theme via replacement (replacing letters with cliparts); however our replacement is for the purposes of lexical resolution, not visual. 
Recently, there has been an exploration of neural style transfer for logo generation \cite{atarsaikhan2018contained}. 
This work however only transfers color and texture from the style to the content. 
Unlike our approach, it does not alter the shape. 
GANvas Studio~\footnote{\url{https://ganvas.studio/}} is a creative venture that uses Generative Adverserial Networks (GANs) \cite{goodfellow2014generative} to create abstract paintings. 
GANs have also been used for logo generation \cite{sage2018logo}.

Recently Google's QuickDraw! and AutoDraw based on sketch-rnn~\cite{ha2018neural} have gained a lot popularity. 
Their work trains a recurrent neural network (RNN) to construct stroke-based drawings of common objects, and is also able to recognize objects from human strokes. 
One could envision creating a doodle by writing out one letter at a time, that AutoDraw would match to the closest object in its library. 
However, these matches would not be theme based. 
Iconary~\footnote{\url{https://iconary.allenai.org/}} is a very recent pictionary-like game that users can play with an AI. 
Relevant to this work, user drawings in Iconary are mapped to icons from The Noun Project to create a scene.

The use of conditional adversarial networks for image-to-image translation is gaining popularity. 
However, using a pix2pix-like architecture~\cite{isola2017image} for our task would involve the use of labeled pairwise (clipart, letter) data, which as discussed earlier, is hard to obtain. 
CycleGAN~\cite{zhu2017unpaired} does not require paired label data, but is not a good fit for our task because we are interested in matching letters to cliparts from a specific theme. 
The pool of clipart is thus limited, and would not be sufficient to learn the target domain. 
Finally, generative modeling is typically lossy; we prefer direct replacement of cliparts for greater readability.

\subsection{Creativity through visual appearance of text}
Advances in conditional GANs~\cite{mirza2014conditional} have motivated style transfer for fonts~\cite{azadi2018multi} through few-shot learning. 
Work on learning a manifold of fonts~\cite{campbell2014learning} allows everyday users to create and edit fonts by smoothly interpolating between existing fonts. 
The former explores the creation of unseen letters of a known font, and the latter explores the creation of entirely new fonts -- neither add any theme-related semantics or additional graphic elements to the text.

Work on neural font style transfer between fonts~\cite{atarsaikhan2017neural} explores the effects of using different weighted factors, character placements and orientations in style transfer. 
This work also has an experiment using icons as style images, however the style transfer is only within the context of visual features of icons such as the texture and thickness of strokes, as opposed to direct replacement.

An interesting direction for expressing emotions through text has been explored through papers on colorful text~\cite{mohammad2011colourful},~\cite{kawakami2016character}. 
These works attempt to learn a word-color association. 
Their results show that even though abstract concepts or sequences of characters may not be physically visualizable, they tend to have color associations which can be realized.

MarkMaker~\footnote{\url{https://emblemmatic.org/markmaker/}} generates logos based on company names -- primarily displaying the name in various fonts and styles, sometimes along with a clipart. 
It uses a genetic algorithm to iteratively refine suggestions based on user feedback.

\section{Approach}
\label{sec:approach}

\begin{figure}[t]
\centering
\includegraphics[scale=0.78]{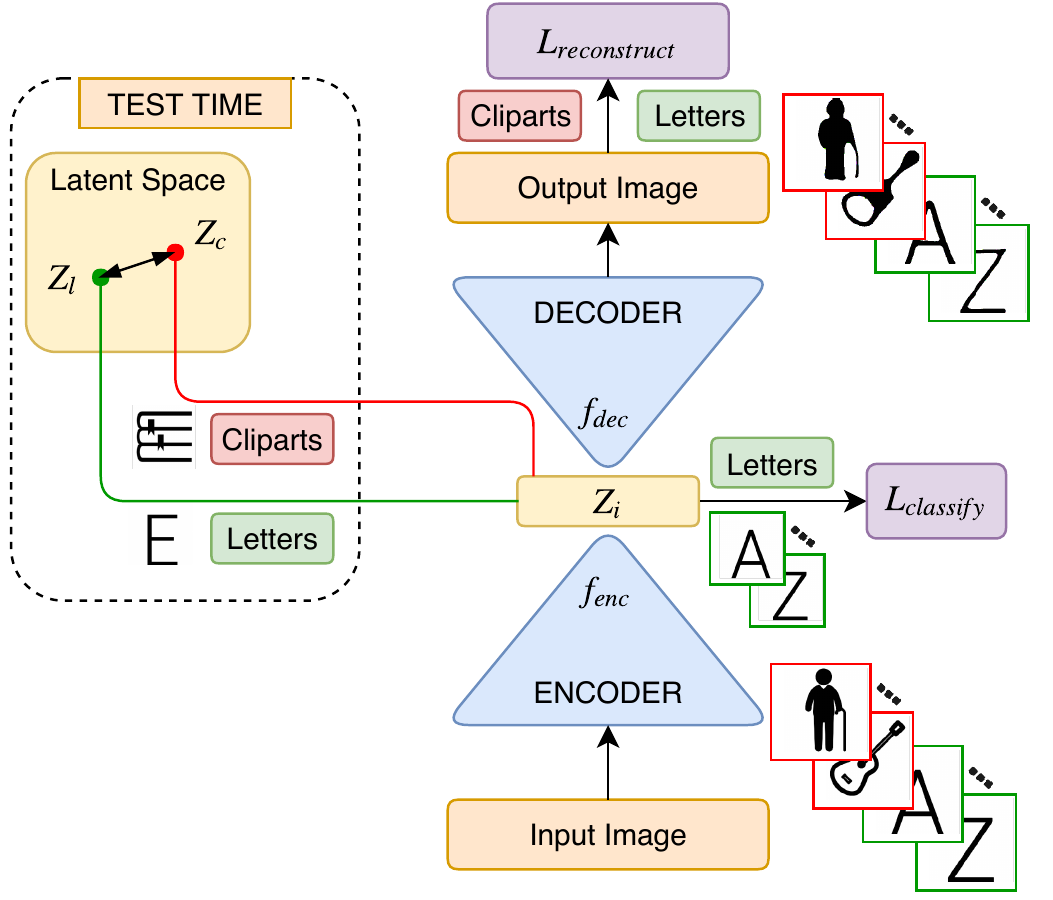}
\caption{Block diagram describing our model during training and testing. The model is trained on a reconstruction and classification loss in a multitask fashion. 
During inference, latent space distances are calculated to match letters to cliparts. See text for more details.}
\label{fig:model}
\end{figure}

\begin{figure*}[t]
	\centering
	\begin{subfigure}{0.5\columnwidth}
		\centering
		\includegraphics[width=0.82\columnwidth]{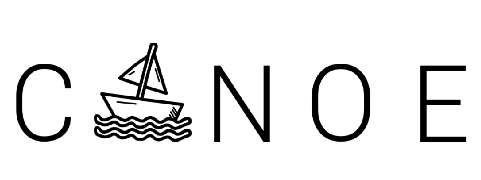}
		\caption{}
	\end{subfigure}
	\begin{subfigure}{0.5\columnwidth}
		\centering
        \includegraphics[width=0.82\columnwidth]{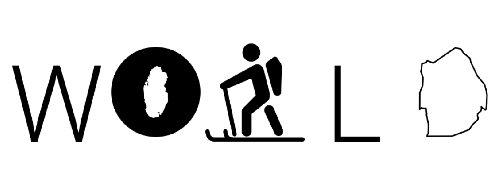}
        \caption{}
	\end{subfigure}
	\begin{subfigure}{0.5\columnwidth}
		\centering
		\includegraphics[width=0.82\columnwidth]{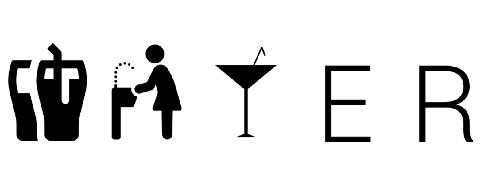}
		\caption{}
	\end{subfigure}
	\begin{subfigure}{0.5\columnwidth}
		\centering
        \includegraphics[width=0.82\columnwidth]{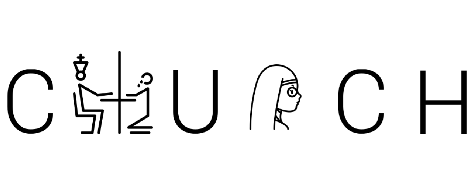}
        \caption{}
	\end{subfigure}
	\vspace{9pt}
	\caption{Example \treat s generated by our approach for (word \& theme) pairs: a) (\code{canoe} \& \code{watersports}) b) (\code{world} \& \code{countries, continents, natural wonders}) c) (\code{water} \& \code{drinks}) d) (\code{church} \& \code{priest, nun, bishop})}
	\vspace{5pt}
    \label{fig:example_outputs}
\end{figure*}

\begin{figure*}[t]
	\centering
	\begin{subfigure}{0.48\columnwidth}
		\centering
		\includegraphics[width=0.85\columnwidth]{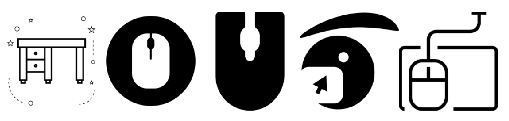}
		\themeall
	\end{subfigure}
	\begin{subfigure}{0.48\columnwidth}
		\centering
        \includegraphics[width=0.85\columnwidth]{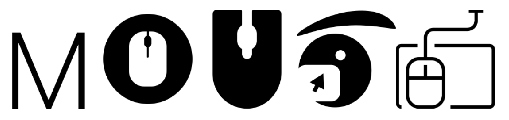}
        \themesome
	\end{subfigure}
	\rulesep
	\begin{subfigure}{0.48\columnwidth}
		\centering
		\includegraphics[width=0.85\columnwidth]{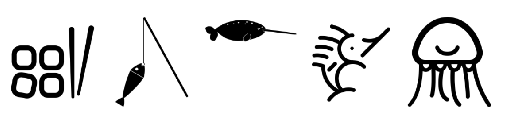}
		\themeall
	\end{subfigure}
	\begin{subfigure}{0.48\columnwidth}
		\centering
        \includegraphics[width=0.85\columnwidth]{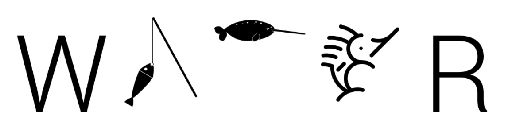}
        \themesome
	\end{subfigure}
	\vspace{3pt}
	\caption{We replace letters in a word with cliparts only if the clipart is sufficiently similar to the letter, placing more stringent conditions on the first and last letters in the word. Notice that in each pair, the \treat s on the right (with a subset of letters replaced) are more legible (\code{Mouse} and \code{Water}) than the ones on the left, while still depicting the associated themes (\code{computer} and \code{fish, mermaid, sailor}).}
    \label{fig:replace_subset}
\end{figure*}

In this section, we first describe our procedure for collecting training data, and then our model and its training details. Finally, we describe our test-time procedure to generate a \treat, that is, obtaining theme-based clipart matches for an input word and theme. A sketch of our model along with examples for training and testing are shown in \figureref{model}. 

\subsection{Training Data}
For our task we need two types of data for training -- letters in different fonts, and cliparts. 
Note that we do not need a \emph{correspondence} between the letters and cliparts. 
In that sense, as stated earlier, our approach is an unsupervised one.

For clipart data, we use the Noun Project -- a website that aggregates and categorizes symbols that are created and uploaded by graphic designers around the world. 
The Noun Project cliparts are all 200 $\times$ 200 in PNG format. 
The Noun Project has binary cliparts, and will result in \treat s of the style shown in \figureref{exam}. 
Different choices of the source of cliparts can result in different styles, including colored \treat s as shown in \figureref{replacement}. 
We downloaded a random set of $\sim$50k cliparts from the Noun Project.

We obtain our letter data from a collection of 1400 distinct font files. \footnote{These font files (TTF) were obtained from a designer colleague.} 
On manual inspection, we found that this set contained a lot of visual redundancies (e.g. the same font being repeated in regular and bold weight types). 
We removed such repetitions. 
We also manually inspected the data to ensure that the individual letters were recognizable in isolation, and discarded overly complicated and intricate font styles. 
This left us with a total of 777 distinct fonts. 
We generated 200 $\times$ 200 image files (PNG format) from each font file for the entire alphabet (uppercase and lowercase) giving us a total of 40.4k images of letters (777 fonts $\times$ 26 letters in the English alphabet $\times$ 2 (upper and lower cases)).

\subsection{Model}
Our primary objective is to find visual similarities between cliparts and letters in an unsupervised manner. 
To this end, we train an autoencoder~\cite{ballard1987modular} with a reconstruction loss on both clipart and letter images (denoted by $\mathbf{X}_{cl}$). 
We denote a single input image by $X_{i}$. 
Each input image $X_{i}$ is passed through an encoder neural network $f_{\text{enc}}(\cdot)$ and projected to a low dimensional intermediate representation $Z_i$. 
Finally, a decoder neural network $f_{\text{dec}}(\cdot)$ tries to reconstruct the input image as $\hat{X_{i}}$, using the objective $\mathcal{L}_{\text{reconstruct}}$,

\begin{align*}
Z_i &= f_{\text{enc}}(X_i) \numberthis \label{eqn1} \\
\hat{X_{i}} &= f_{\text{dec}}(Z_i)  \numberthis \label{eqn_use_better_labels} \\
\mathcal{L}_{\text{reconstruct}} &= \frac{1}{|\mathbf{X}_{cl}|} \sum_{i \in \mathbf{X}_{cl}}{\text{SSD}(X_i,\hat{X_i})} 
\numberthis \label{eqn2}
\end{align*}

where $\text{SSD}(X_i,\hat{X_i})$ is the sum over squared pixel differences between the original image and its reconstruction. 
We set the dimensionality of $Z_i$ to be $128$. 
In addition to the reconstruction objective, we utilize letter labels (52 labels for lowercase and uppercase letters) to classify the intermediate representations $Z_i$ for the letter images. 
This objective helps the encoder discriminate between different letters (possibly with similar visual features) while clustering together the intermediate representations for the same letter across different fonts. 
This would allow the intermediate representation to capture visual features that are characteristic of each letter, and when cliparts are matched to letters using this representation, the matched cliparts will retain the visually discriminative features of letters. 

Concretely, we project $Z_i$ to a 52-dimensional space using a single linear layer with a softmax non-linearity and use the cross entropy loss function. 
Let $W$ and $b$ be the parameters of a linear transformation of $Z_{i}$. 
We obtain a probability distribution $P_i(\cdot)$ across all labels as,

\begin{align*}
P_i(\cdot) &= \text{softmax}(WZ_{i} + b) \numberthis \label{eqn3}
\end{align*}
Let $\mathbf{X}_l$ be the subset of images in $\mathbf{X}_{cl}$ that are letters. We maximize the probability of the correct label $Y_i$ corresponding to each letter image $X_i$. 
\begin{align*}
\mathcal{L}_\text{classify} &= -\frac{1}{|\mathbf{X}_{l}|} \sum_{i \in \mathbf{X}_{l}}{\log{P_{i}(Y_i)}} \numberthis \label{eqn4}
\end{align*}

Note that the same $Z_i$ is used in both objective functions for letter images. These objectives are jointly trained using a multitask objective
\begin{align*}
\mathcal{L} &= \alpha \mathcal{L}_\text{reconstruct} + (1 - \alpha) \mathcal{L}_\text{classify} \numberthis \label{eqn5}
\end{align*}

Our final loss function is thus composed of two different loss functions: (1) $\mathcal{L}_\text{reconstruct}$ trained on both letters and cliparts, and (2) $\mathcal{L}_\text{classify}$ trained only on letters. 
Here $\alpha$ is a tunable hyperparameter in the range $[0, 1]$. 
We set $\alpha$ to 0.25 after manually inspecting outputs of a few word-theme pairs we used while developing our model (different from the word-theme pairs we use to evaluate our approach later).

\subsubsection{Implementation Details:}
Our encoder network $f_{\text{enc}}(\cdot)$ is an AlexNet~\cite{krizhevsky2012imagenet} convolutional neural network trained from scratch, made up of 5 convolutional and 3 fully connected layers. 
Our decoder network $f_{\text{dec}}$ consists of 5 deconvolutional layers, 3 fully connected layers and 3 upsampling layers. We use batch norm between layers.\footnote{Implementations of our encoder and decoder were adapted from \url{https://github.com/arnaghosh/Auto-Encoder}.} 
We use ReLU activations for both the encoder and decoder. 
We use the Adam optimizer with a learning rate of $10^{-4}$, and a weight decay of $10^{-5}$. 
The input dataset is divided into minibatches of size 100 with a mixture of clipart and letter images in each minibatch. 
We use early stopping based on a validation set as our stopping criterion.

\subsubsection{Data Preprocessing:}
We resize our images to 224 $\times$ 224 using bilinear interpolation to match the input size of our AlexNet-based encoder. 
We normalize every channel of our input data to fall in $[-1, 1]$.

\subsection{Finding Matches}
At test time given a word  and a theme, we retrieve a theme-relevant pool of cliparts  (denoted by $\mathbf{C}$) by querying Noun Project. 
If multiple phrases have been used to describe a theme, we use each phrase separately as a query. 
We limit this retrieval to no more than 10,000 cliparts for each phrase. 
We then combine cliparts for different phrases of a theme together to form the final pool of cliparts for that theme. 
For example, for the theme \code{countries, continents, natural wonders}, we query Noun Project for \code{countries}, \code{continents} and \code{natural wonders} individually and combine all retrieved cliparts together to form the final pool of theme-relevant cliparts. 
On average across 95 themes we experimented with, we had a minimum of 49 and maximum of 29,580 cliparts per theme, with a mean of 9731.2 and median of 9966. 
We augmented this set of cliparts with left-right (mirror) flips of the cliparts. 
This improves the overall match quality. E.g., in cases where there existed a good match for \code{J}, but not for \code{L}, the clipart match for \code{J}, when flipped, served as a good match for \code{L}. Similarly for \code{S} and \code{Z}.

For each letter $l$ of the input word, we choose the corresponding letter images (denoted by $\mathbf{F}_l$) taken from a predefined pool of fonts. 
To create the pool of letter images $\mathbf{F}_l$, we used uppercase letters from 14 distinct, readable fonts from among the 777 fonts used during training. 
These were kept fixed for all experiments. 
We found that uppercase letters had better matches with the cliparts (lower cosine distance between corresponding latent representations $Z_i$ on average, and visually better matches). 
Moreover, we found that in several fonts, letters were the same for both cases.

We replace the letter with a clipart (chosen from $\mathbf{C}$) whose mean cosine distance in the intermediate latent space is the least, when computed against every letter image in $\mathbf{F}_l$. 
Concretely, if $Z_i$ denotes the intermediate representation for the $i^{th}$ image in $\mathbf{F}_l$ and $Z_c$ is the intermediate representation for a clipart $c$ in $\mathbf{C}$, the chosen clipart $\hat{c}_l$ is

\begin{align*}
\hat{c}_l = \argmin_{c \in \mathbf{C}} \frac{1}{|\mathbf{F}_l|} \sum_{i \in \mathbf{F}_l}{\left( 1 - \frac{Z_{i} \cdot Z_{c}}{|Z_i||Z_c|}\right)} \numberthis \label{eqn6}
\end{align*}

We find a clipart that is most similar to the letter on average across fonts to ensure a more robust match than considering a single most similar font. 
In this way, each letter in the input word is replaced by its closest clipart to generate a \treat.

We show example \treat  s generated by our approach in \figureref{example_outputs}. We find that the word can often be difficult to recognize from the \treat~if the Noun Project cliparts corresponding to a theme are not sufficiently similar to the letters in the word. 
To improve the legibility of our \treat s, we first normalize the cosine distance values of our matched cliparts for the alphabet for a specific theme in the range [0, 1]. 
We only replace a letter with its clipart match if the normalized cosine distance between the embedding of the letter and clipart is $<$ 0.75. 
It is known that the first and last letters of a word play a crucial role in whether humans can recognize the word at a glance. 
So we use a stricter threshold, and replace the first and last letters of a word with a clipart only if the normalized cosine distance between the two is $<$ 0.45. 
Example \treat s with all letters replaced and only a subset of letters replaced can be seen in \figureref{replace_subset}. 
Clearly, the \treat s with a subset of letters replaced (\themesome) are more legible than replacing all letters (\themeall), while still depicting the desired theme. 
We quantitatively evaluate this in the next section.

\section{Evaluation}
\label{sec:evaluation}

\begin{figure*}[t]
	\centering
	\begin{subfigure}{1\columnwidth}
		\centering
		\includegraphics[width=0.63\columnwidth]{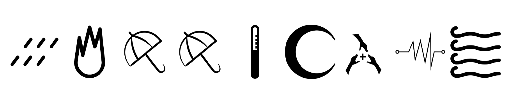}
		\caption*{\themeall}
		\vspace{15pt}
	\end{subfigure}
	\begin{subfigure}{1\columnwidth}
		\centering
        \includegraphics[width=0.6\columnwidth]{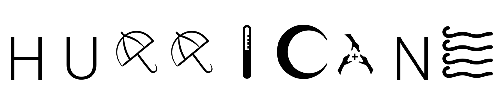}
        \caption*{\themesome}
        \vspace{15pt}
	\end{subfigure}
	\begin{subfigure}{0.66\columnwidth}
		\centering
		\includegraphics[width=0.9\columnwidth]{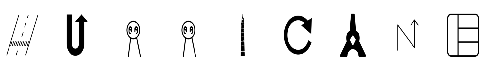}
		\vspace{3pt}
		\caption*{\nothemeall}
		\vspace{15pt}
	\end{subfigure}
	\begin{subfigure}{0.66\columnwidth}
		\centering
        \includegraphics[width=0.9\columnwidth]{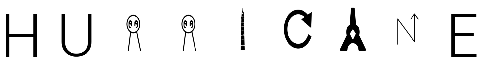}
        \vspace{3pt}
        \caption*{\nothemesome}
        \vspace{15pt}
	\end{subfigure}
	\begin{subfigure}{0.66\columnwidth}
		\centering
        \includegraphics[width=0.82\columnwidth]{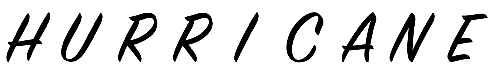}
        \vspace{3pt}
        \caption*{\fontall}
        \vspace{15pt}
	\end{subfigure}
	\caption{We evaluate five different approaches for generating \treat s. See text for details.}
    \label{fig:all_doodle_types}
\end{figure*}

\begin{figure}[t]
\centering
\includegraphics[width=0.9\columnwidth]{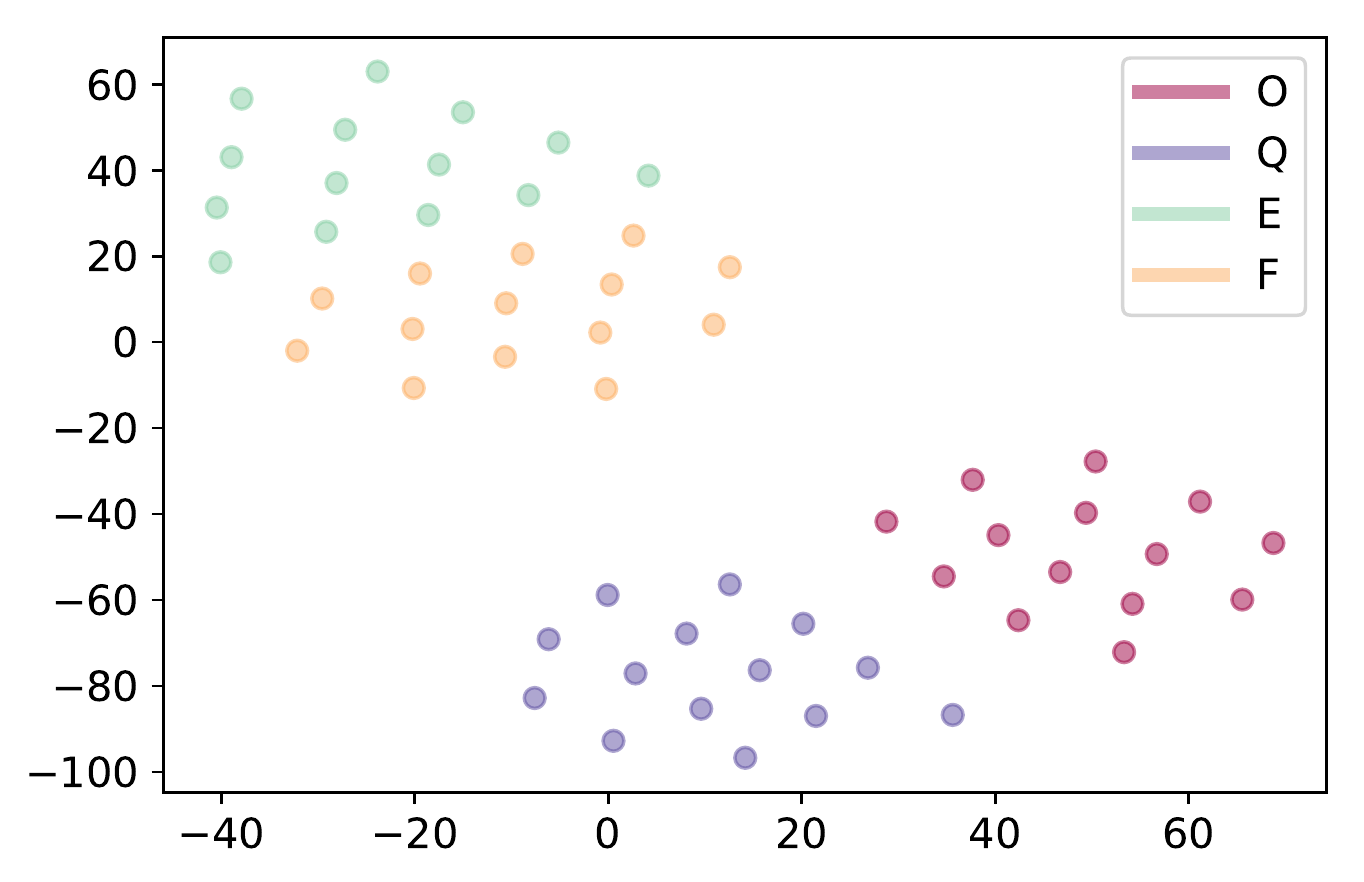}
\caption{t-SNE plot showing clusters of uppercase \code{O}, \code{Q}, \code{E} and \code{F}. Each letter forms its own cluster and visually similar pairs (\code{E} \& \code{F}, \code{O} \& \code{Q}) form super-clusters. However, these super-clusters are far apart from each other due to significant visual differences.}
\label{fig:oqef}
\end{figure}

\begin{figure}[t]
\centering
\includegraphics[width=1\columnwidth]{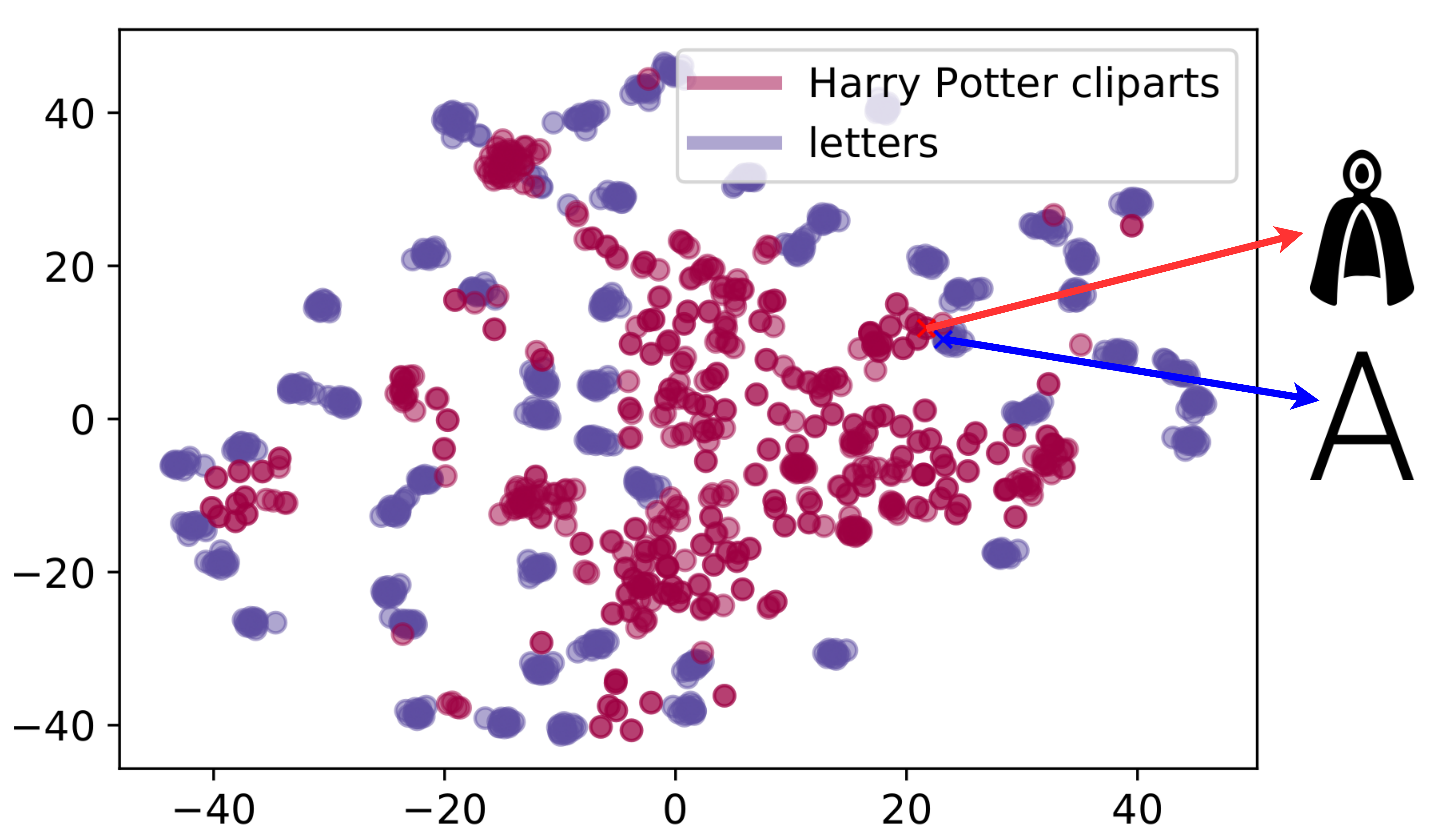}
\caption{t-SNE plot showing clusters of \code{Harry Potter} themed cliparts along with letters. Cliparts which look like \code{A} lie close to the cluster of \code{A}'s in the latent space.}
\label{fig:hp_a}
\end{figure}

\begin{figure}[t]
	\centering
	\begin{subfigure}{0.48\columnwidth}
		\centering
		\includegraphics[width=1\columnwidth]{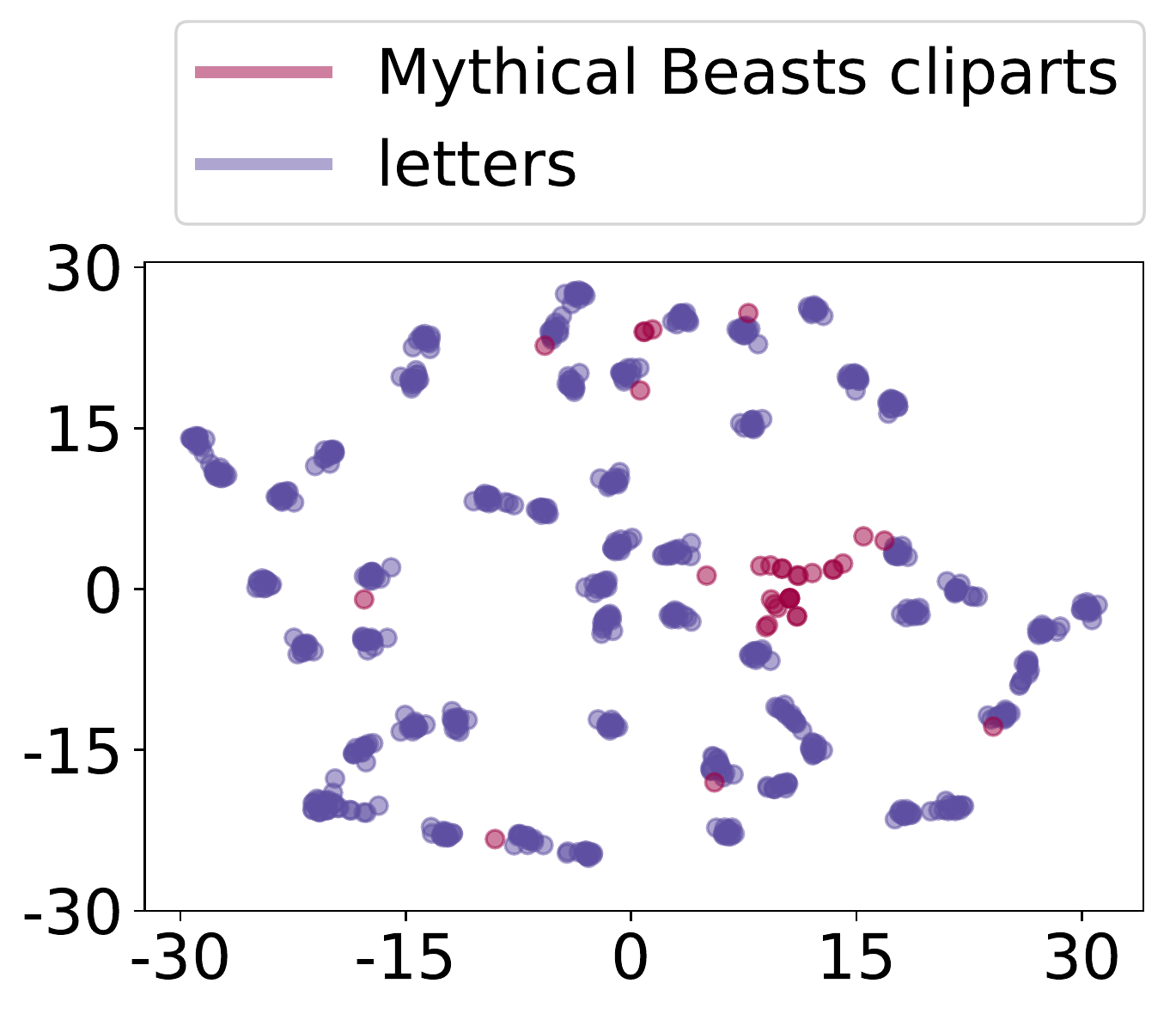}
		\vspace{4pt}
		\includegraphics[width=1\columnwidth]{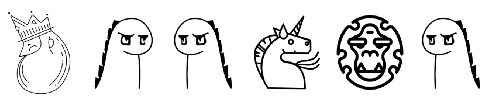}
		\caption{}
		\label{fig:dragon}
		\vspace{8pt}
	\end{subfigure}%
	\begin{subfigure}{0.52\columnwidth}
		\centering
        \includegraphics[width=1\columnwidth]{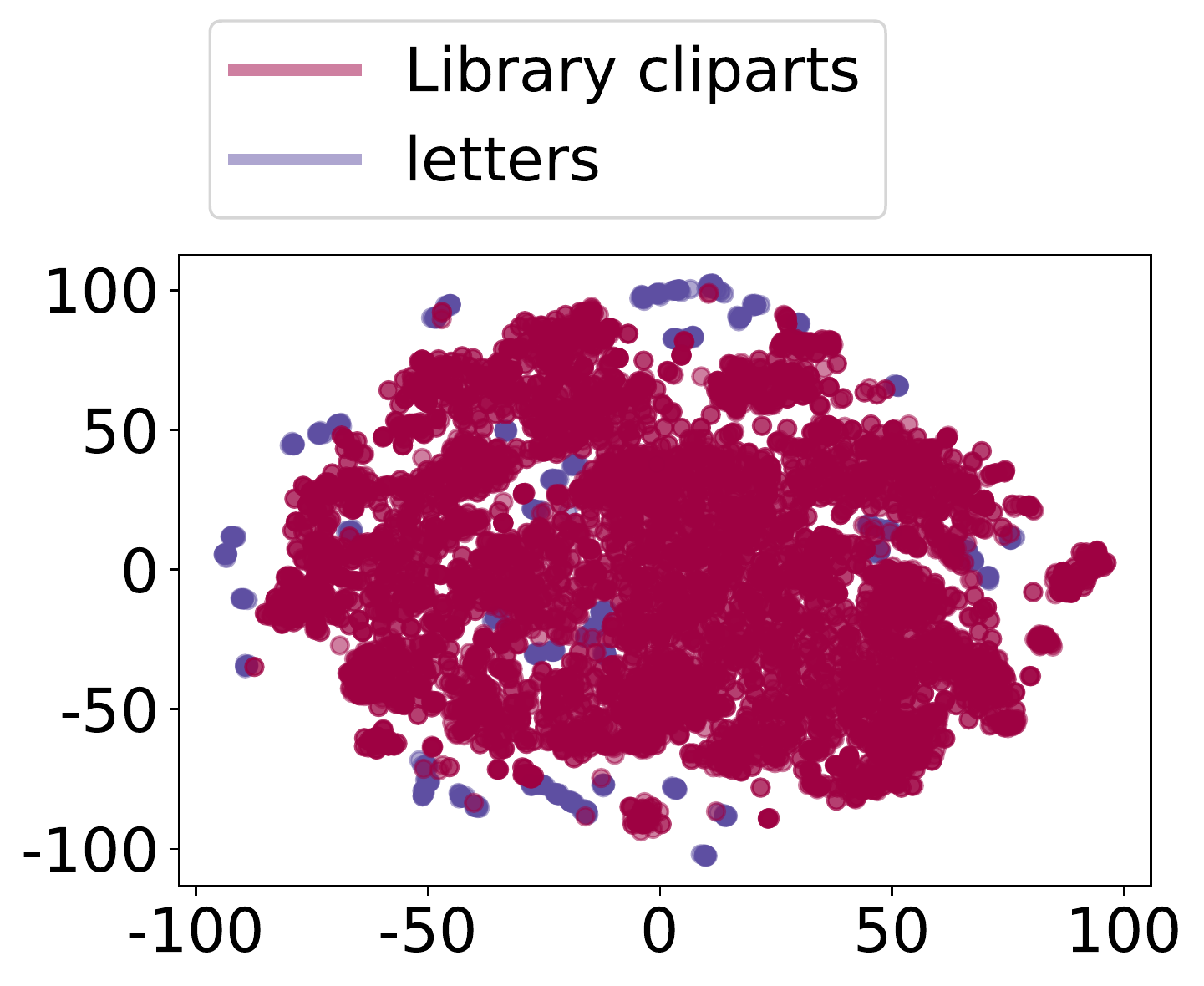}
        \includegraphics[width=0.7\columnwidth]{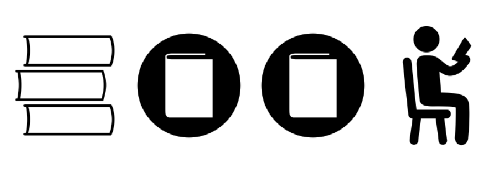}
        \caption{}
        \label{fig:book}
        \vspace{8pt}
	\end{subfigure}

	\caption{Impact of diversity of cliparts from different themes in the Noun Project on corresponding \treat s. a) \treat~of \code{dragon} in theme \code{mythical beast} is not legible due to lower coverage of letters by the themed cliparts compared to a \treat~of \code{book} in theme \code{library} shown in b).}
    \label{fig:diversity}
\end{figure}

\begin{figure*}[t]
	\centering
	\begin{subfigure}{0.35\columnwidth}
	    \centering
		\includegraphics[width=0.99\columnwidth]{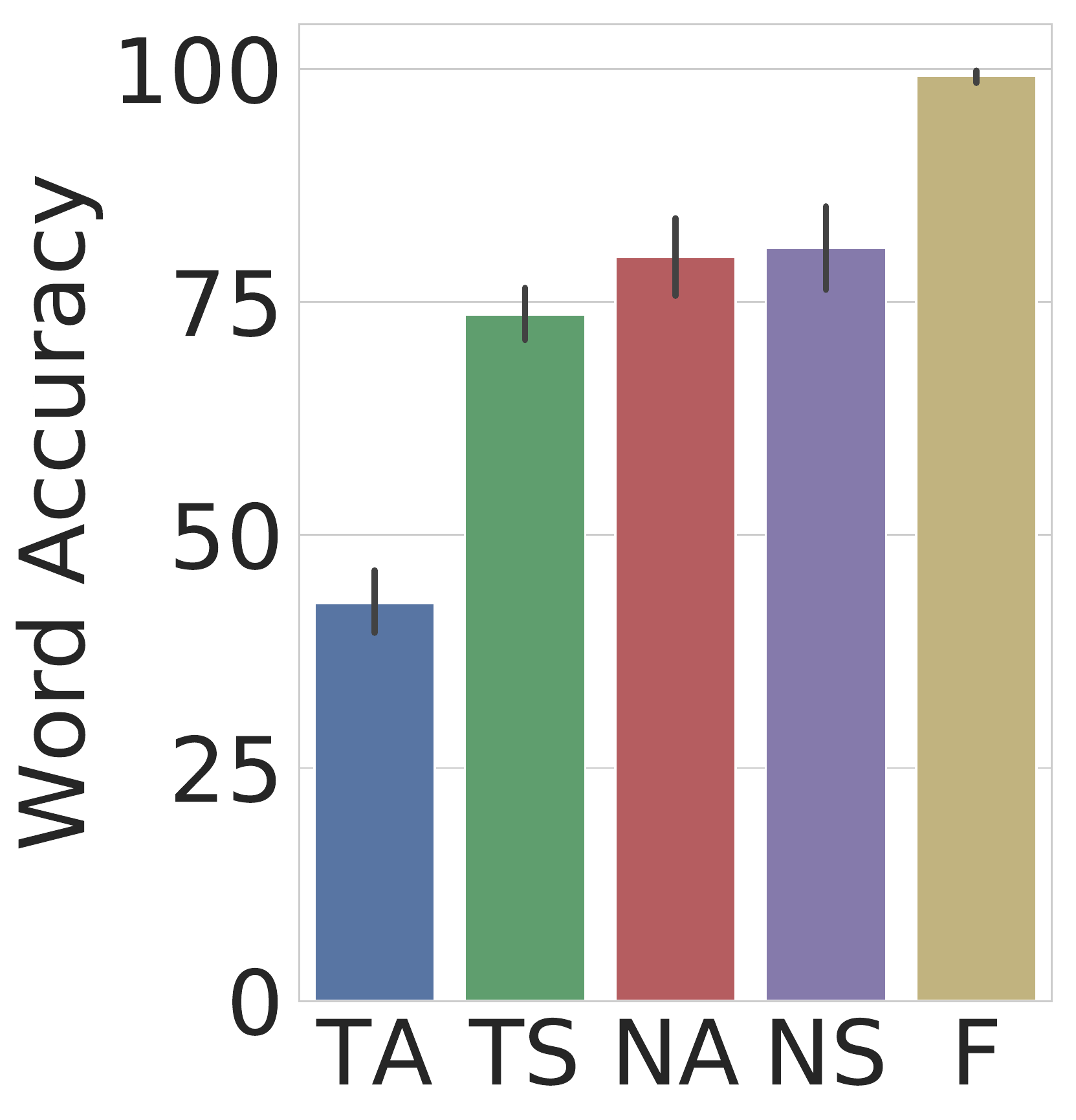}
		\caption{}
		\label{fig:word_acc}
	\end{subfigure}
	\begin{subfigure}{0.35\columnwidth}
		\centering
        \includegraphics[width=0.99\columnwidth]{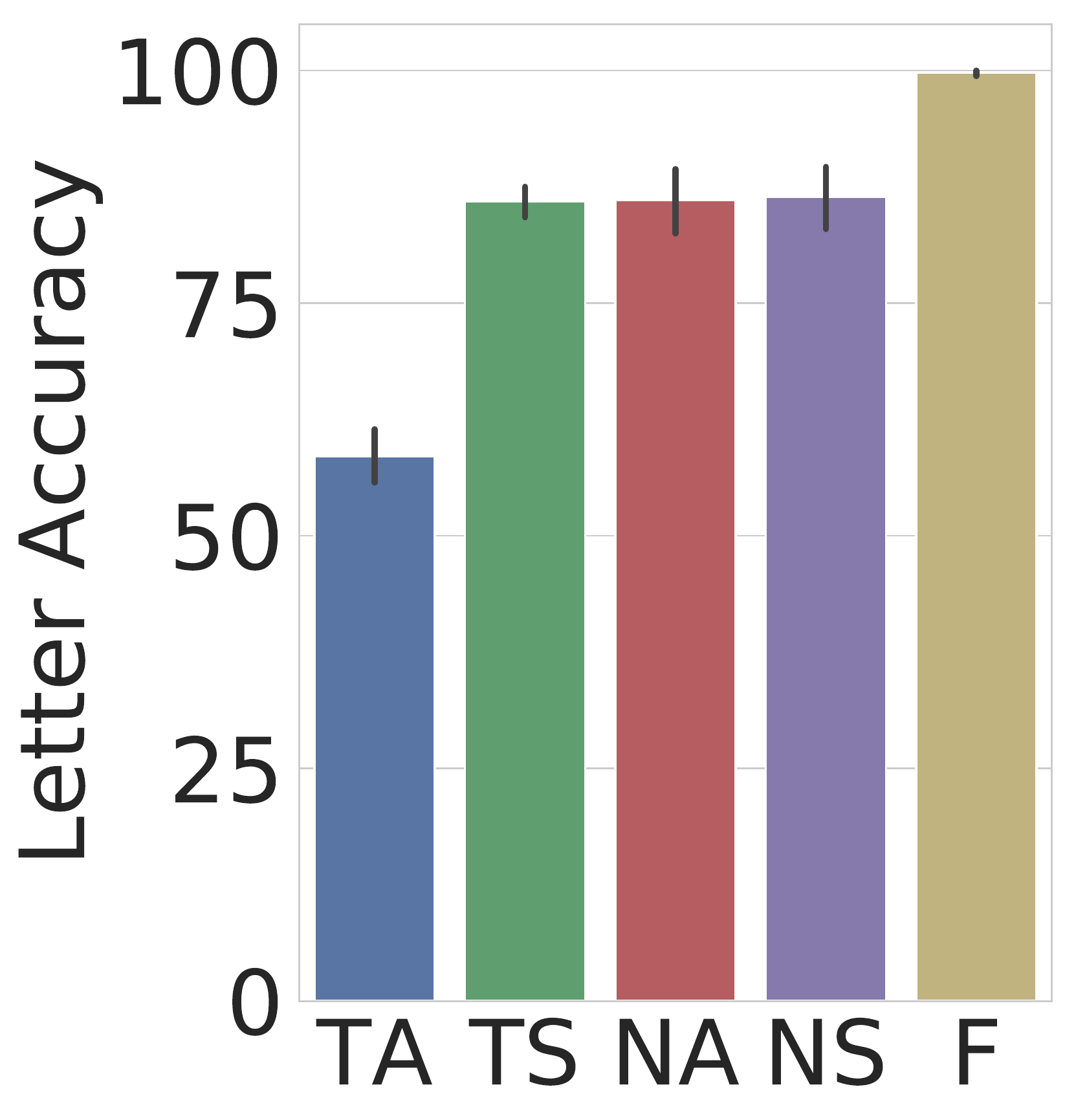}
        \caption{}
        \label{fig:letter_acc}
	\end{subfigure}
	\begin{subfigure}{0.35\columnwidth}
	    \centering
		\includegraphics[width=0.99\columnwidth]{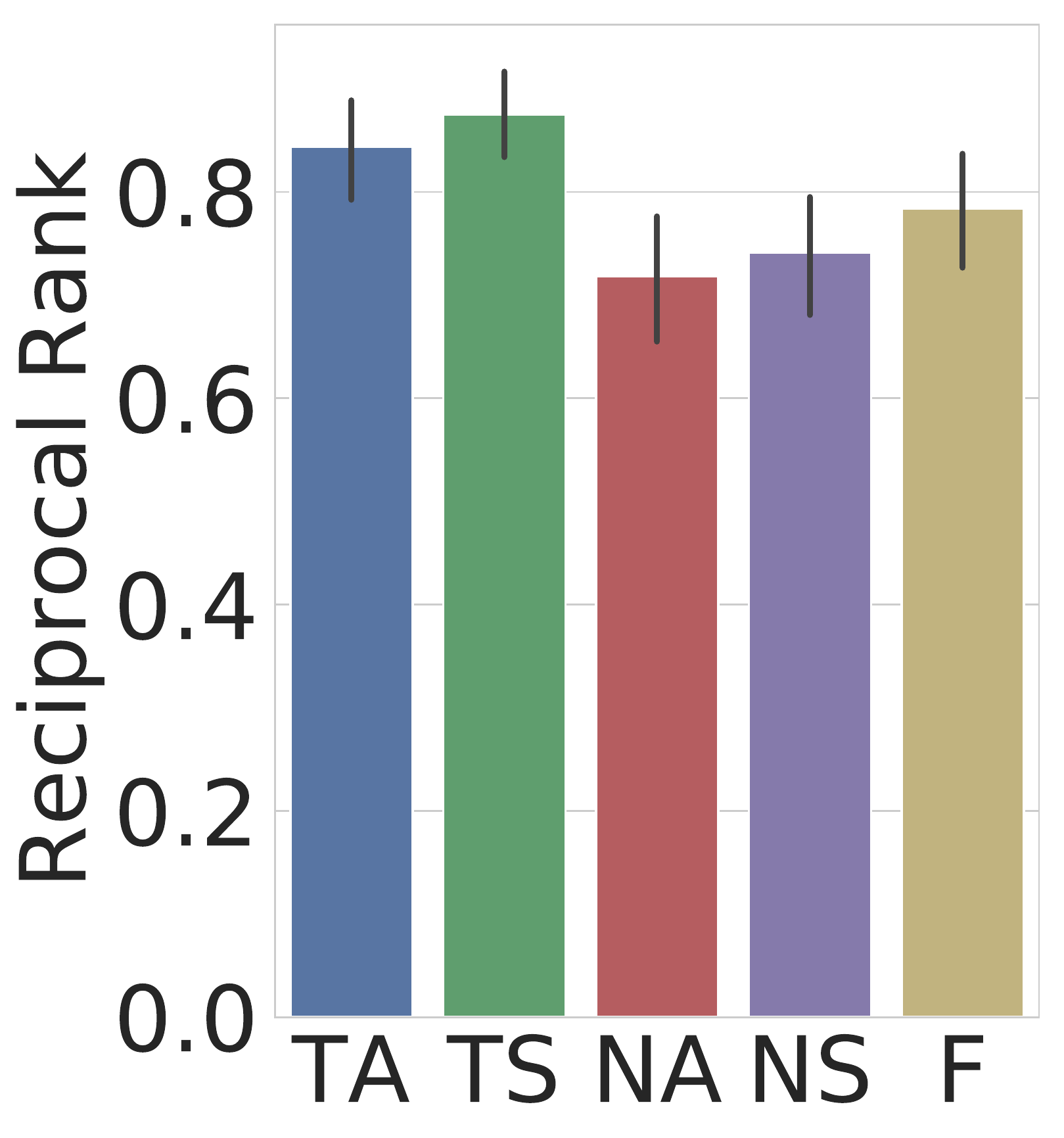}
		\caption{}
		\label{fig:mrr}
	\end{subfigure}
	\begin{subfigure}{0.62\columnwidth}
		\centering
        \includegraphics[width=0.99\columnwidth]{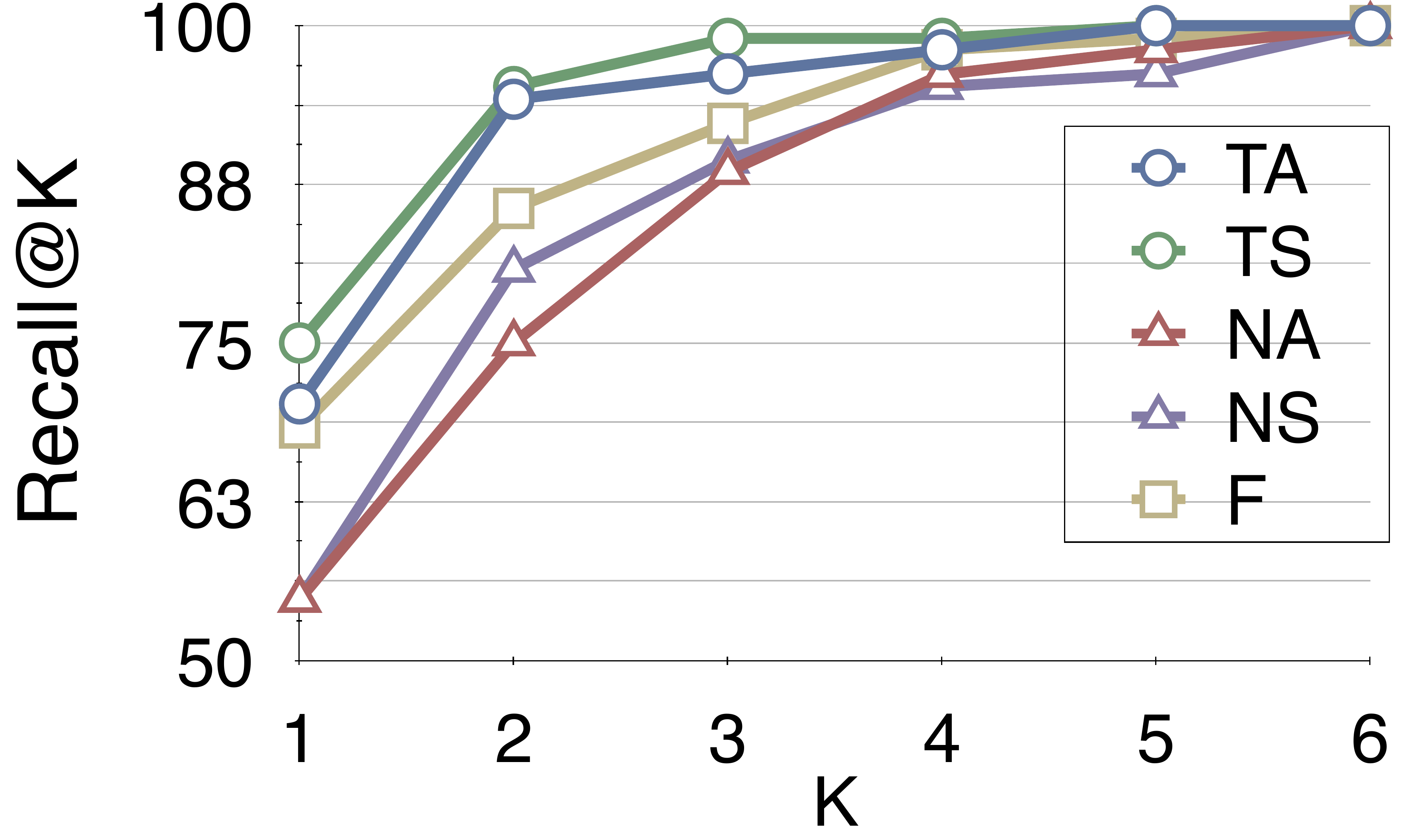}
        \caption{}
        \label{fig:recall}
	\end{subfigure}
	\begin{subfigure}{0.33\columnwidth}
		\centering
        \includegraphics[width=0.99\columnwidth]{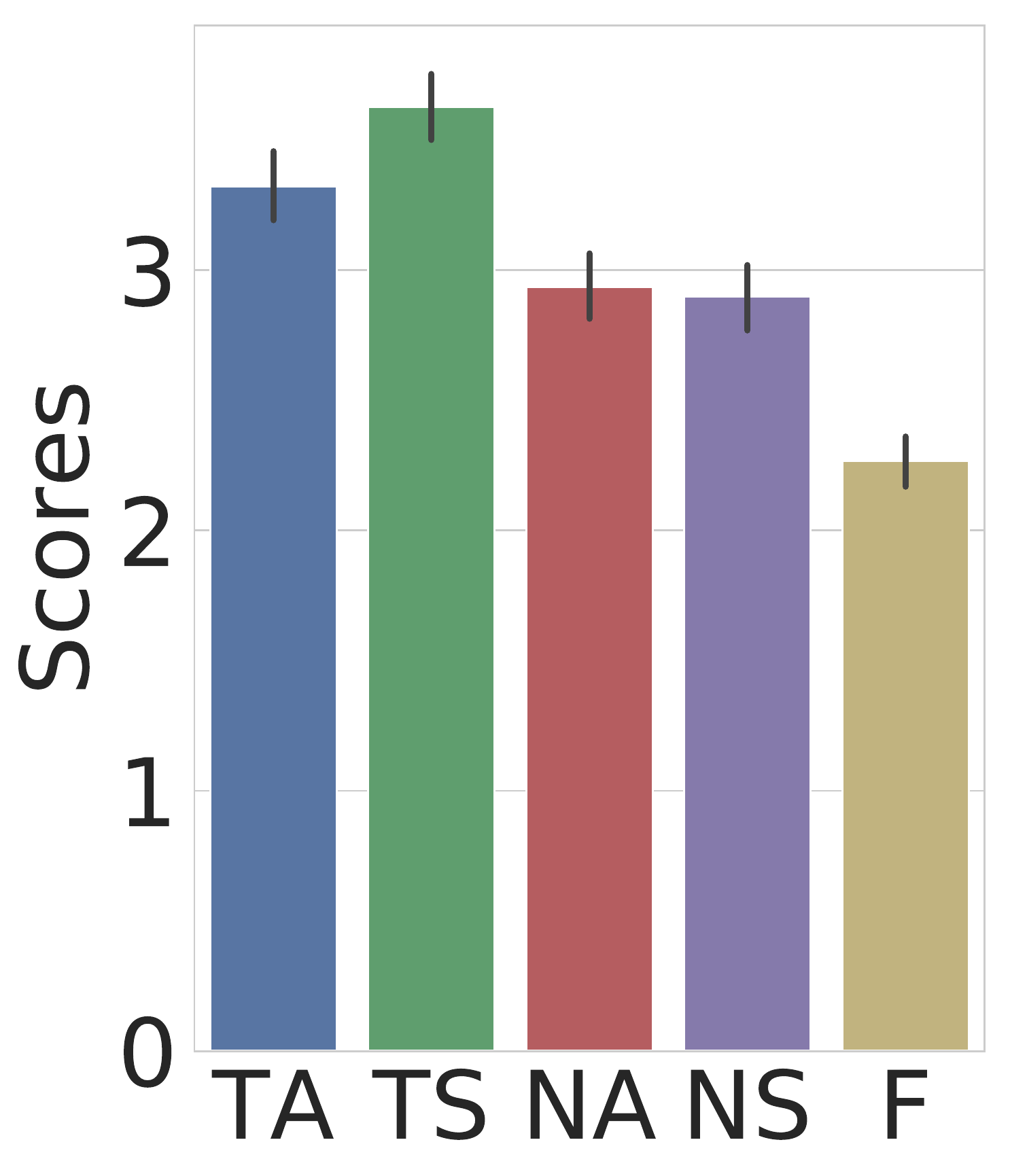}
        \caption{}
        \label{fig:creativity}
	\end{subfigure}
	\vspace{6pt}
	\caption{Evaluation of \treat s from five approaches (\themeall (TA), \themesome (TS), \nothemeall (NA), \nothemesome (NS), \fontall (F))
	for a) word recognition; b) letter recognition; c) and d) theme recognition; e) creativity.
	}
    \label{fig:word-compare2}
\end{figure*}

We evaluate our entire system along three dimensions:
\begin{itemize}
    \item How well is our model able to learn a representation that captures visual features of the letters?
    \item How does our chosen source of cliparts (Noun Project) affect the quality of matches?
    \item How good are our generated \treat s?
\end{itemize}

\subsection{Learnt Representation}
We use t-Distributed Stochastic Neighbor Embedding (t-SNE) to visualize our learnt latent representations of letters and cliparts. 
Among letters, we find that our model clusters letters of different fonts together, while  distinguishing between visually dissimilar letters. 
E.g., \figureref{oqef}  visualizes in 2D uppercase \code{O}, \code{Q}, \code{E} and \code{F} in the 14 fonts used at test time. 
As expected, \code{O} and \code{Q} clusters are close, and \code{E} and \code{F} clusters are close, but both these sets of clusters are apart. Visualizing letters as well as cliparts, 
\figureref{hp_a} shows that our model is able to learn a representation such that visually similar letter-clipart pairs are close in the latent space.

\subsection{Effect of source of cliparts}
Themes which have fewer cliparts, and hence lower diversity and coverage across the letters (e.g. \code{mythical beast} in \figureref{dragon}) have poorer matches as compared to larger, more diverse themes (e.g. \code{library} in \figureref{book}). 
Indeed, we see that recognizing the word in a \treat~generated from the former theme is significantly harder than for the latter.

\subsection{Quality of \treat s}
We now evaluate the quality of \treat s generated by our approach. 
We developed our approach on a few themes (e.g., \code{education}, \code{Harry Potter}, \code{Halloween}, \code{Olympics}) and associated words (e.g., \code{exam}, \code{always}, \code{witch}, \code{play}). 
To evaluate our approach in the open world, we collected 104 word-theme pairs from subjects on Amazon Mechanical Turk (AMT). 
We told subjects that given a word and an associated theme, we have a bot that can draw a doodle. 
We showed subjects a few example \treat s. 
We asked subjects to give us a word and an associated theme (to be described in 1-5 comma separated phrases) that they would like to see a doodle for. 
Example (word \& theme) pairs from our dataset are (\code{environment} \& \code{pollution, dirt, wastage}), (\code{border} \& \code{USA}), (\code{computer} \& \code{technology}). 
We allowed subjects to use multiple phrases to describe the theme to allow for a more diverse set of cliparts to search from when generating the \treat. 
We evaluate our \treat s along three dimensions:
\begin{itemize}
    \item Can subjects recognize the word in the \treat ?
    \item Can subjects recognize the theme in the \treat ?
    \item Do subjects find the \treat~creative?
\end{itemize}

We compare our approach \themesome~to a version where we replace all letters in the word with cliparts (\themeall) to evaluate how replacing a subset of letters affects word recognition (expected to increase) and theme recognition (expected to remain unchanged or even increase because recognizing the word can aid in recognizing the theme), as well as creativity (expected to remain unchanged or even increase because the associated word is more legible as opposed to gibberish). 
We also compare our approach to an approach that replaces letters with cliparts, but is not constrained by the theme of interest (\nothemesome~and \nothemeall). 
We find the clipart that is closest across all 95\footnote{95 because some themes repeat in our 104 (word \& theme) pairs.} themes in our dataset to replace the letter. 
This can result in increased word recognition because letters can find a clipart that is more similar (from a larger pool not constrained by the theme), but will result in lower theme recognition accuracy. 
Note that theme recognition will still likely be higher than chance because the word itself gives cues about the theme. 
For no-themed clipart, we compare an approach that replaces all letters (i.e., \nothemeall) as well as only a subset of letters (\nothemesome). 
Finally, as a point of reference, we evaluate a \treat~that simply displays the word in a slightly atypical font (\fontall). 
We expect word recognition to be nearly perfect, but theme recognition as well as creativity to be poor. 
These five different types of \treat s are shown in \figureref{all_doodle_types}. 
This gives us a total of 520 \treat s to evaluate (5 types $\times$ 104 word-theme input pairs). 
No AMT workers were repeated across any of these tasks.

\subsubsection{Word recognition:}

We showed each \treat~to 5 subjects on AMT. 
They were asked to type out the word they see in the \treat~in free-form text. 
Notice the open-ended nature of the task. 
Performance of crowd-workers for word recognition of different types of \treat s is shown in \figureref{word_acc}.
This checks for exact string matching (case-insensitive) between the word entered by subjects and the true word. 
As a less stringent evaluation, we also compute individual letter recognition accuracy. 
These were computed only for cases where the length of the word entered by the subject matched the true length of the \treat~because if the lengths do not match, the worker likely made a mistake or was distracted. 
Letter recognition accuracies are shown in \figureref{letter_acc}.

As expected, leaving a subset of the letters unchanged leads to a higher recognition rate for \themesome~and \nothemesome~compared to their counterparts, \themeall~and \nothemeall~respectively. 
Also, \nothemeall~and \nothemesome~have higher word recognition accuracy than \themeall~and \themesome~because the clipart matches are obtained from a larger pool (across all themes rather than from a specific theme). 
The added signal from the theme of the cliparts in \themeall~and \themesome~does not help word recognition enough to counter this. 
\nothemeall~already has a high recognition rate, leaving little scope for improvement for \nothemesome. 
Finally, \fontall~has near perfect word recognition accuracy because it contains the word clearly written out. 
It is not a 100\% because of typos on the part of the subjects.  
In some cases we found that subjects did not read the instructions and wrote out the theme instead of the word itself across all \treat s. 
These subjects were excluded from our analysis. 

\subsubsection{Theme recognition:}

We showed each \treat~to 6 subjects on AMT. 
The same theme can be described in many different ways. 
So unlike word recognition, this task could not be open-ended. 
For each \treat , we gave subjects 6 themes as options from which the correct theme is to be identified. 
These 6 options included the true theme from the 95 themes in our dataset, 2 similar themes, and 3 random themes. 
The similar themes are the 2 nearest neighbor themes to the true theme in \texttt{word2vec}~\cite{mikolov2013distributed} space. 
\texttt{word2vec} is a popular technique to generate vector representations of a word or ``word embeddings'' which capture the meaning of the word such that words that share common contexts in language (that is, likely have similar meaning) are located in close proximity to one another in the space. 
If a theme is described by multiple words, we represent the theme using the average \texttt{word2vec} embedding of each word. 
This is a strategy that is commonly employed in natural language processing to reason about similarities between phrases or even entire sentences~\cite{wieting2016towards},~\cite{adi2017fine}. 
We find that 64\% of the \treat s were assigned to the correct theme for \themeall, 67\% for \themesome, 43\% for \nothemeall, 51\% for \nothemesome~and 60\% for \fontall~respectively. 
As expected, \nothemeall~and \nothemesome~have lower theme recognition accuracy than \themeall~and \themesome~because \nothemeall~and \nothemesome~do not use cliparts from specific themes. 
Notice that theme recognition accuracy is still quite high, because the word itself often gives away cues about the theme (as seen by the theme recognition accuracy of \fontall~that lists the word without any clipart). 

This theme recognition rate is a pessimistic estimate  because theme options presented to subjects included nearest neighbors to the true theme as distractors. 
These themes are often synonymous to the true theme. 
As a less stringent evaluation, we sort the 6 options for each \treat~based on the number of votes the option got across subjects. 
\figureref{mrr} shows the Mean Reciprocal Rank of the true option in this list (higher is better). 
We also show Recall@K in \figureref{recall} that compute how often the true option is in the top-K in this sorted list. 
Similar trends as described above hold.

Comparing \themesome~to \themeall, we see that replacing only a subset of letters does not hurt theme recognition (in fact, it improves slightly), but improves word recognition significantly. So overall, \themesome~produces the best \treat s. 
We see this being played out when \treat s are evaluated for their overall creativity (next). 
This relates to Schmidhuber's theory of creativity~\cite{schmidhuber2010formal}. 
He argues that data is creative if it exhibits both a learnable or recognizable pattern (and is hence compressible), and novelty. 
\themesome~achieves this balance.

\subsubsection{Creativity:}
Recall that our goal here is to create \treat s to depict words with visual elements such that the \treat~leaves an impression on people's minds. 
We now attempt to evaluate this. Do subjects find the \treat~intriguing / surprising / fun (i.e., creative)?
We showed each \treat~to 5 subjects on AMT. 
They were told: ``This is a doodle of \code{[word]} in a \code{[theme]} theme. 
On a scale of 1-5, how much do you agree with this statement? 
This doodle is creative (i.e, surprising and/or intriguing and/or fun). 
1. Strongly agree (with a grin-like smiley face emoji in green) 
2. Somewhat agree (with a smiley face in lime green) 
3. Neutral (with a neutral face in yellow) 
4. Somewhat disagree (with a slightly frowning face in orange) 
5. Strongly disagree (with a frowning face in red).'' 
Crowd-worker ratings are shown in \figureref{creativity}. 
\themesome~was rated the highest. 
We believe this is due to a good trade off between legibility and having a theme-relevant depiction that allows for semantic reinforcement. 
\nothemeall~and \nothemesome~are significantly worse. 
Recall that they are visual, but not in a theme-specific way. 
So they are visually interesting, but do not allow for semantic reinforcement. 
The resultant reduction in creativity is evident. 
Interestingly, \nothemesome~scores slightly higher than \nothemeall. 
This may be because \nothemesome~is not more legible than \nothemeall~(\nothemeall~is already sufficiently legible). 
With more of the letters visually depicted, \nothemeall~is more interesting. Finally,  \fontall~has a significantly lower creativity score. 
It is rated lower than neutral, close to the ``Somewhat disagree'' rating. 
To get a qualitative sense, we asked subjects to comment on what they think of the \treat s. 
Some example comments: \\
\themeall: \textit{``cool characters and each one fits the theme of the ocean''}, \textit{``Its [sic] creative and represents the theme well, but I don't see disney all that much.''} \\
\themesome: \textit{``I like how it uses the image of the US and then a state to spell out the word and looks like something you'd remember.''}, \textit{``Very fun and intriguing. I like how all the letters are pictures representing a computer mouse.''}. \\
\nothemeall: \textit{``It is creative but it has nothing to do with fear.''}, \textit{``It does a very good job of spelling out CHRISTMAS, but the individual letters are not related to the holiday at all.''} \\ 
\nothemesome: \textit{``It is somewhat creative, especially the unicorn head for ``G'', though I don't know what any of it has to do with the theme.''}, \textit{``There are too many icons that seemingly have nothing to do with the theme.''} \\
\fontall: \textit{``It just spells out the word, not really a doodle''}, \textit{``Its not a doodle, its just the word Parrot, so I don't think its creative at all.''}

\begin{figure*}[ht!]
	\centering
	\begin{subfigure}{0.5\columnwidth}
		\centering
		\includegraphics[width=\columnwidth]{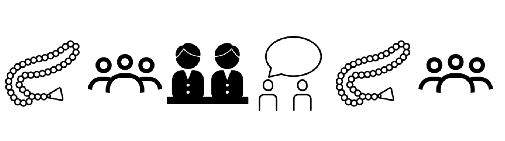}
		\caption{}
		\label{fig:church}
		\vspace{8pt}
	\end{subfigure}%
	\rulesep
	\begin{subfigure}{0.5\columnwidth}
		\centering
        \includegraphics[width=1\columnwidth]{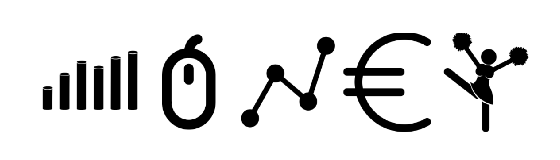}
        \caption{}
        \label{fig:money}
        \vspace{8pt}
	\end{subfigure}
	\rulesep
	\begin{subfigure}{0.5\columnwidth}
		\centering
        \includegraphics[width=0.95\columnwidth]{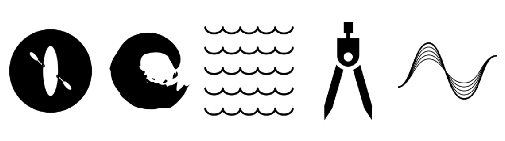}
        \themeall
        \label{fig:ocean_d1}
        \vspace{8pt}
	\end{subfigure}
		\begin{subfigure}{0.5\columnwidth}
		\centering
        \includegraphics[width=0.95\columnwidth]{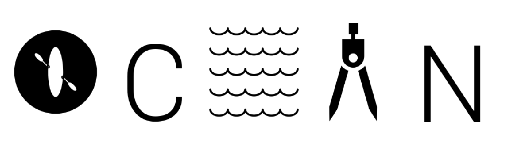}
        \themesome
        \label{fig:ocean_d2}
        \vspace{8pt}
	\end{subfigure}

	\caption{Example failure modes of our approach. See text for details.}
    \label{fig:all_bad}
\end{figure*}

\section{Generative Morphing}
\label{sec:generative}

\begin{figure}[t]
	\centering
	\begin{subfigure}{0.17\columnwidth}
		\centering
		\includegraphics[width=\columnwidth]{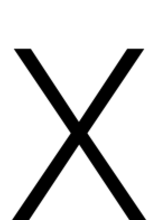}
		letter
	\end{subfigure}
	\rulesep
	\begin{subfigure}{0.57\columnwidth}
	    \centering
    		\includegraphics[width=0.3\columnwidth]{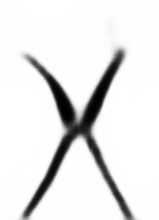}
    		\includegraphics[width=0.3\columnwidth]{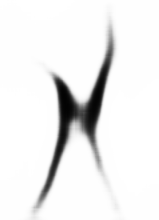}
    		\includegraphics[width=0.3\columnwidth]{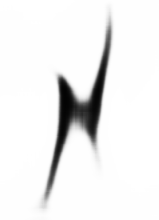}
    	morphs
    \end{subfigure}
    \rulesep
	\begin{subfigure}{0.17\columnwidth}
		\centering
        \includegraphics[width=\columnwidth]{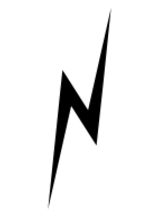}
        clipart
	\end{subfigure}
	\vspace{3pt}
	\caption{The letter \code{X} (left) and its clipart match (scar) in the theme \code{Harry Potter} (right) with three generated morphs in between. The morphs are modified versions of the scar so that it looks more like the \code{X} while still being recognizable as Harry Potter's scar.}
    \label{fig:generative}
\end{figure}

Recall that our model has an autoencoder as a component. 
The decoder takes as input the latent embedding and generates an image corresponding to it. 
This presents us with an opportunity to explore morphing mechanisms to make the \treat s more interesting. 
We can start with the embedding corresponding to a letter and smoothly interpolate to the embedding corresponding to the matched clipart. 
At each step along the way, we can generate an image that depicts the intermediate visual. 
This allows for increased legibility (because the original letter is visible early in the morph), as well as potential for more semantic reinforcement and intrigue as the \treat~is slowly ``revealed'' over time. 
An example of a morphed output is shown in \figureref{generative}. 
The matched clipart doesn't naturally resemble the letter \code{X} as much as it does the letter \code{N}. 
Morphing allows for the clipart to be transformed in a way that makes the letter (\code{X}) more apparent, while still retaining its visual identity (\code{Harry Potter}'s scar).

\section{Future Work}
\label{sec:future_work}

In this section, we discuss some drawbacks of our current model and potential future work.

\subsubsection{No Clipart Relevance Score:}

A comment from a subject evaluating the creativity of \figureref{church} (word \code{church} \& theme \code{pastor, Jesus, people, steeple}) was \textit{``you need a cross [...] before the general public would [...] get this.''} 
Our approach does not include world knowledge that indicates which symbols are canonical for themes (Noun Project does not provide a relevance score). 
As a result, our model can not explicitly trade off visual similarity (VS) for theme relevance (TR) -- either to compromise on VS to improve TR, or to at least optimize for TR if VS is poor.

\subsubsection{No Contextual Querying:}
Multiple phrases used to describe a theme often lose context when they are used individually to query Noun Project. 
For example, the last clipart in \figureref{money} for (word \code{money} \& theme \code{finance, banking, support}) is of a cheerleader, and hence relevant to the phrase \code{support}, but is not relevant in the context of the \code{finance} theme.

The lack of context also hurts polysemous theme words. 
\code{bat} when used as a keyword with the another keyword \code{bird} refers to the creature bat, but in the context of \code{baseball} refers to sports equipment.

\subsubsection{Imperfect Match Scores:}
Our automatic similarity score frequently disagrees with our (human) perceptual notion of similarity. 
E.g., in \figureref{all_bad} right, the cliparts used to replace \code{C} and \code{N} in \themeall~look sufficiently similar to the corresponding letters. 
But the automatic similarity score was low, and so \themesome~chose to not replace the letters. 
Approaches to improve the automatic score can be explored in future work. 
For instance, in addition to mirror images, using rotated and scaled versions of the cliparts to augment the dataset would help.

\subsection{Interactive Interface}
To mitigate these concerns, we plan to build an interactive tool. 
Users can choose from the top-$k$ clipart matches for each letter. 
Users can iterate on the input theme descriptions until they are satisfied with the \treat. 
Users can also leave the theme unspecified in which case we can use the word itself as the theme. 
Finally, users can choose which letters to replace in \themesome~like \treat s.

\section{Conclusion}
\label{sec:conclusion}

In this work, we introduce a computational approach for semantic reinforcement called \treat~-- Thematic Reinforcement for Artistic Typography. 
Given an input word and a theme, our model generates a ``doodle'' (\treat) for that word using cliparts associated with that theme. 
We evaluate our \treat s for word recognition (can a subject recognize the word being depicted?), theme recognition (can a subject recognize what theme is being illustrated in the \treat ?), and creativity (overall, do subjects find the \treat s surprising / intriguing / fun?). 
We find that subjects can recognize the word in our \treat s 74\% of the time, can recognize the theme 67\% of the time, and on average ``Somewhat agree'' that our \treat s are creative.

\section{Acknowledgements}
\label{sec:acknowledgements}

We thank Abhishek Das, Harsh Agrawal, Prithvijit Chattopadhyay and Karan Desai for their valuable feedback.

\balance{}

\bibliographystyle{iccc}
\bibliography{iccc}

\begin{thebibliography}{}

\bibitem[\protect\citeauthoryear{Adi \bgroup et al.\egroup
  }{2017}]{adi2017fine}
Adi, Y.; Kermany, E.; Belinkov, Y.; Lavi, O.; and Goldberg, Y.
\newblock 2017.
\newblock Fine-grained analysis of sentence embeddings using auxiliary
  prediction tasks.
\newblock In {\em ICLR}.

\bibitem[\protect\citeauthoryear{Atarsaikhan \bgroup et al.\egroup
  }{2017}]{atarsaikhan2017neural}
Atarsaikhan, G.; Iwana, B.~K.; Narusawa, A.; Yanai, K.; and Uchida, S.
\newblock 2017.
\newblock Neural font style transfer.
\newblock In {\em ICDAR}.

\bibitem[\protect\citeauthoryear{Atarsaikhan, Iwana, and
  Uchida}{2018}]{atarsaikhan2018contained}
Atarsaikhan, G.; Iwana, B.~K.; and Uchida, S.
\newblock 2018.
\newblock Contained neural style transfer for decorated logo generation.
\newblock In {\em IAPR DAS}.

\bibitem[\protect\citeauthoryear{Azadi \bgroup et al.\egroup
  }{2018}]{azadi2018multi}
Azadi, S.; Fisher, M.; Kim, V.~G.; Wang, Z.; Shechtman, E.; and Darrell, T.
\newblock 2018.
\newblock Multi-content gan for few-shot font style transfer.
\newblock In {\em CVPR}.

\bibitem[\protect\citeauthoryear{Ballard}{1987}]{ballard1987modular}
Ballard, D.~H.
\newblock 1987.
\newblock Modular learning in neural networks.
\newblock In {\em AAAI}.

\bibitem[\protect\citeauthoryear{Campbell and
  Kautz}{2014}]{campbell2014learning}
Campbell, N.~D., and Kautz, J.
\newblock 2014.
\newblock Learning a manifold of fonts.
\newblock {\em Trans. on Graphics}.

\bibitem[\protect\citeauthoryear{Clawson \bgroup et al.\egroup
  }{2012}]{clawson2012using}
Clawson, T.~H.; Leafman, J.; Nehrenz~Sr, G.~M.; and Kimmer, S.
\newblock 2012.
\newblock Using pictograms for communication.
\newblock {\em Military medicine}.

\bibitem[\protect\citeauthoryear{Frutiger}{1989}]{frutiger1989signs}
Frutiger, A.
\newblock 1989.
\newblock Signs and symbols.
\newblock {\em Their design and meaning}.

\bibitem[\protect\citeauthoryear{Goodfellow \bgroup et al.\egroup
  }{2014}]{goodfellow2014generative}
Goodfellow, I.; Pouget-Abadie, J.; Mirza, M.; Xu, B.; Warde-Farley, D.; Ozair,
  S.; Courville, A.; and Bengio, Y.
\newblock 2014.
\newblock Generative adversarial nets.
\newblock In {\em NIPS}.

\bibitem[\protect\citeauthoryear{Ha and Eck}{2018}]{ha2018neural}
Ha, D., and Eck, D.
\newblock 2018.
\newblock A neural representation of sketch drawings.
\newblock In {\em ICLR}.

\bibitem[\protect\citeauthoryear{Isola \bgroup et al.\egroup
  }{2017}]{isola2017image}
Isola, P.; Zhu, J.-Y.; Zhou, T.; and Efros, A.~A.
\newblock 2017.
\newblock Image-to-image translation with conditional adversarial networks.
\newblock In {\em CVPR}.

\bibitem[\protect\citeauthoryear{Kawakami \bgroup et al.\egroup
  }{2016}]{kawakami2016character}
Kawakami, K.; Dyer, C.; Routledge, B.; and Smith, N.~A.
\newblock 2016.
\newblock Character sequence models for colorful words.
\newblock In {\em EMNLP}.

\bibitem[\protect\citeauthoryear{Krizhevsky, Sutskever, and
  Hinton}{2012}]{krizhevsky2012imagenet}
Krizhevsky, A.; Sutskever, I.; and Hinton, G.~E.
\newblock 2012.
\newblock Imagenet classification with deep convolutional neural networks.
\newblock In {\em NIPS}.

\bibitem[\protect\citeauthoryear{Martins \bgroup et al.\egroup
  }{2015}]{martins2015good}
Martins, P.; Urbancic, T.; Pollak, S.; Lavrac, N.; and Cardoso, A.
\newblock 2015.
\newblock The good, the bad, and the aha! blends.
\newblock In {\em ICCC}.

\bibitem[\protect\citeauthoryear{Martins, Cunha, and
  Machado}{2018}]{martins2018how}
Martins, P.; Cunha, J.~M.; and Machado, P.
\newblock 2018.
\newblock How shell and horn make a unicorn: Experimenting with visual blending
  in emoji.
\newblock In {\em ICCC}.

\bibitem[\protect\citeauthoryear{Mikolov \bgroup et al.\egroup
  }{2013}]{mikolov2013distributed}
Mikolov, T.; Sutskever, I.; Chen, K.; Corrado, G.~S.; and Dean, J.
\newblock 2013.
\newblock Distributed representations of words and phrases and their
  compositionality.
\newblock In {\em NIPS}.

\bibitem[\protect\citeauthoryear{Mirza and
  Osindero}{2014}]{mirza2014conditional}
Mirza, M., and Osindero, S.
\newblock 2014.
\newblock Conditional generative adversarial nets.
\newblock {\em arXiv preprint arXiv:1411.1784}.

\bibitem[\protect\citeauthoryear{Mohammad}{2011}]{mohammad2011colourful}
Mohammad, S.
\newblock 2011.
\newblock Colourful language: Measuring word-colour associations.
\newblock In {\em Workshop on Cognitive Modeling and Computational Linguistics,
  ACL}.

\bibitem[\protect\citeauthoryear{Sage \bgroup et al.\egroup
  }{2018}]{sage2018logo}
Sage, A.; Agustsson, E.; Timofte, R.; and Van~Gool, L.
\newblock 2018.
\newblock Logo synthesis and manipulation with clustered generative adversarial
  networks.
\newblock In {\em CVPR}.

\bibitem[\protect\citeauthoryear{Schmandt-Besserat}{2015}]{schmandt2014evolution}
Schmandt-Besserat, D.
\newblock 2015.
\newblock The evolution of writing.
\newblock {\em International Encyclopedia of the Social and Behavioral
  Sciences: Second Edition}.

\bibitem[\protect\citeauthoryear{Schmidhuber}{2010}]{schmidhuber2010formal}
Schmidhuber, J.
\newblock 2010.
\newblock Formal theory of creativity, fun, and intrinsic motivation
  (1990--2010).
\newblock {\em IEEE Transactions on Autonomous Mental Development}.

\bibitem[\protect\citeauthoryear{Shiojiri and
  Nakatani}{2013}]{shiojiri2013visual}
Shiojiri, M., and Nakatani, Y.
\newblock 2013.
\newblock Visual language communication system with multiple pictograms
  converted from weblog texts.
\newblock In {\em IASDR}.

\bibitem[\protect\citeauthoryear{Takasaki and Mori}{2007}]{takasaki2007design}
Takasaki, T., and Mori, Y.
\newblock 2007.
\newblock Design and development of a pictogram communication system for
  children around the world.
\newblock In {\em Intercultural collaboration}.

\bibitem[\protect\citeauthoryear{Wieting \bgroup et al.\egroup
  }{2016}]{wieting2016towards}
Wieting, J.; Bansal, M.; Gimpel, K.; and Livescu, K.
\newblock 2016.
\newblock Towards universal paraphrastic sentence embeddings.
\newblock In {\em ICLR}.

\bibitem[\protect\citeauthoryear{Xiao and Linkola}{2015}]{xiaovismantic}
Xiao, P., and Linkola, S.
\newblock 2015.
\newblock Vismantic: Meaning-making with images.
\newblock In {\em ICCC}.

\bibitem[\protect\citeauthoryear{Zhu \bgroup et al.\egroup
  }{2017}]{zhu2017unpaired}
Zhu, J.-Y.; Park, T.; Isola, P.; and Efros, A.~A.
\newblock 2017.
\newblock Unpaired image-to-image translation using cycle-consistent
  adversarial networks.
\newblock In {\em ICCV}.

\end{thebibliography}

\end{document}